\documentclass{article}

 \PassOptionsToPackage{numbers}{natbib}
\usepackage[preprint]{neurips_2026}
\usepackage[utf8]{inputenc}
\usepackage[T1]{fontenc}
\usepackage{graphicx}
\usepackage{float}
\usepackage{subcaption}
\usepackage{microtype}
\usepackage{hyperref}
\usepackage{url}
\usepackage{booktabs}
\usepackage{multicol,multirow,tabularx}
\usepackage{mathtools}
\usepackage{amsmath}
\usepackage{amssymb}
\usepackage{amsfonts}
\usepackage{enumitem}
\usepackage[table]{xcolor}
\usepackage{pgfmath}
\usepackage{arydshln}

\hypersetup{
  colorlinks=false,
  pdfborder={0 0 0}
}

\definecolor{mygreen1}{RGB}{245,250,245}
\definecolor{mygreen2}{RGB}{232,243,232}
\definecolor{mygreen3}{RGB}{214,233,214}
\definecolor{mygreen4}{RGB}{190,221,190}

\definecolor{myred1}{RGB}{250,244,245}
\definecolor{myred2}{RGB}{246,228,232}
\definecolor{myred3}{RGB}{240,209,216}
\definecolor{myred4}{RGB}{233,190,198}

\newcommand{\MidVal}{3.0}
\newcommand{\RedMax}{30.0}

\newcommand{\heatcell}[1]{%
  \begingroup
  \pgfmathparse{#1}%
  \let\val\pgfmathresult
  \ifdim\val pt<\MidVal pt
    \pgfmathparse{(\MidVal-\val)/\MidVal}%
    \let\dist\pgfmathresult
    \ifdim\dist pt<0.12pt
      \cellcolor{mygreen1}#1%
    \else\ifdim\dist pt<0.30pt
      \cellcolor{mygreen2}#1%
    \else\ifdim\dist pt<0.60pt
      \cellcolor{mygreen3}#1%
    \else
      \cellcolor{mygreen4}#1%
    \fi\fi\fi
  \else
    \pgfmathparse{min(1,(#1-\MidVal)/(\RedMax-\MidVal))}%
    \let\dist\pgfmathresult
    \ifdim\dist pt<0.12pt
      \cellcolor{myred1}#1%
    \else\ifdim\dist pt<0.30pt
      \cellcolor{myred2}#1%
    \else\ifdim\dist pt<0.60pt
      \cellcolor{myred3}#1%
    \else
      \cellcolor{myred4}#1%
    \fi\fi\fi
  \fi
  \endgroup
}

\usepackage{amsmath,amsfonts,bm}

\def\eqref#1{equation~\ref{#1}}

\def\1{\bm{1}}

\DeclareMathAlphabet{\mathsfit}{\encodingdefault}{\sfdefault}{m}{sl}
\SetMathAlphabet{\mathsfit}{bold}{\encodingdefault}{\sfdefault}{bx}{n}

\title{\textsc{CustomerSim}: Benchmarking and Aligning Multimodal Language Models as Retail User Simulators}

\author{
  Yada Pruksachatkun\thanks{Equal contribution. Correspondence: \texttt{[ypruksachatkun, wu.jason]@salesforce.com}} \\
  Salesforce Research \\
  \And
  Yixin Wan\footnotemark[1] \\
  University of California Los Angeles \\
  \And
  Xingrun Chen \\
  University of California Los Angeles \\
  \And
  Kai-Wei Chang \\
  University of California Los Angeles \\
  \And
  Chien-Sheng Wu \\
  Salesforce Research \\
}

\begin{document}

\maketitle

\begin{abstract}
We present CustomerSim, an environment and benchmark to evaluate the extent to which Multimodal Large Language Models (MLLMs) can simulate realistic, persona-driven customer behavior in chat-based retail environments. While prior work treats user simulation as surface-level dialog generation, we focus on a model's ability to seek information and make decisions that adhere to customer specifications in multiturn, agentic simulations. CustomerSim consists of a human-curated set of 360 personas over five product categories, alongside a suite of metrics measuring consistency between a customer simulator's actions and its specifications and conversational quality. We find several behavioral gaps across five open and closed-source state-of-the-art models. First, while models produce fluent conversations, they display significantly lower lexical diversity than human shoppers, and open-source models overdisclose their criteria in the opening turn. Second, models tend to be persuaded by sales agent tone and drift from persona specifications. Even the strongest closed-source models, Claude Opus 4.8 and GPT-5.6 Sol, achieves <0.74 alignment with its persona specifications. To address these limitations, we propose UserGRPO, a multi-turn, multi-objective reinforcement learning recipe optimizing both conversational fluency and decision alignment under persona specifications. UserGRPO raises decision alignment of the baseline model from 0.417 to 0.652, a gain of 23.5 points, without meaningful cost to conversational quality, and these gains transfer to held-out product categories. We further find that stylistic prompting is the only intervention that makes surface form more human-like, yet it nearly halves persona adherence. Through CustomerSim, we provide a testbed for the community to investigate and improve the adherence of user simulators in goal-oriented settings.
\end{abstract}
\vspace{-1em}

\section{Introduction}

Large Language Models (LLMs) have been increasingly used as user simulators for economic simulations and evaluation and training of interactive AI systems.~\citep{10.1145/3596510, davidson2023user,hu2023unlocking,kazi2024large,ALGHERAIRY2025101697,zhao2025mua, qian2025userrl, qianuserbench, huang2026crmarenapro,park2023generative,tau2bench}. However, the fidelity of user simulators themselves has been relatively understudied. 
In this work, we study the efficacy of using Multimodal LLMs (MLLM) as customer simulators in retail interactions, inspired by economic simulations from \cite{park2023generative} as well as product usecases such as customer simulators for sales training.
Existing works assess customer simulators on surface-level qualities: \cite{davidson2023user,luo-etal-2024-duetsim} utilize task success rate as the main metric, which is affected by both shopper and sales agent performance. In the retail domain, ~\citet{gromada-etal-2025-evaluating} introduces customer simulator-specific metrics that rely on rating scales and LLM-as-a-judge for persona adherence. Unlike this work, we design our evaluation methodology such that there exists a ground truth mapping between persona specifications and acceptable products in the catalog data, allowing for verifiable evaluation of decision-level persona adherence.


\begin{figure}
\centering
    \includegraphics[width=14cm]{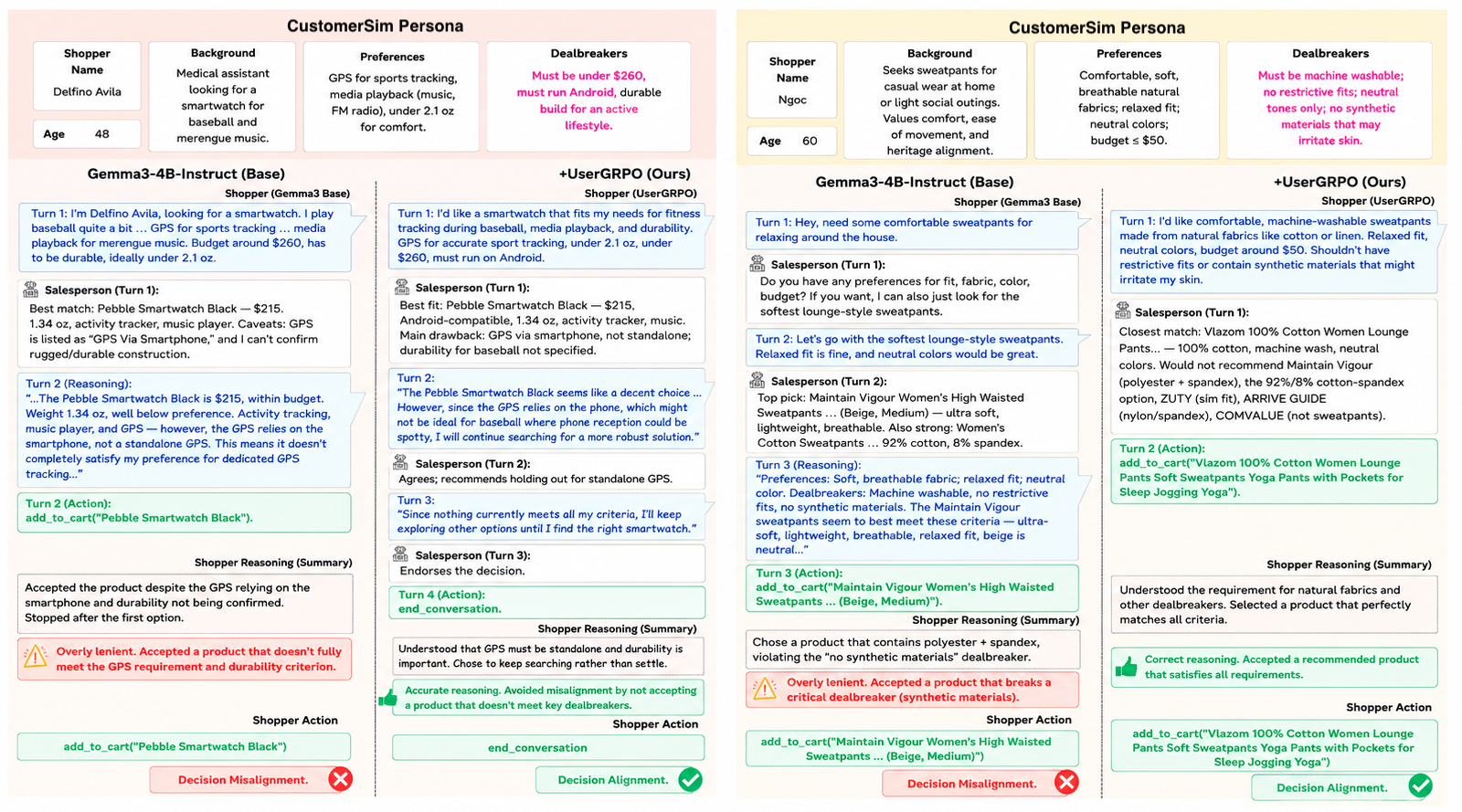}
    \vspace{-1.5em}
    \caption{\textbf{Qualitative examples of retail simulations on \textsc{CustomerSim}.} 
    Baseline models exhibit over-leniency. They are also susceptible to tonality of the salesperson simulator, both in proceeding with unsuitable purchases or rejecting acceptable products as specified by their persona. 
    In contrast, our \textsc{UserGRPO} model demonstrates more grounded reasoning based on product attributes.
    }
\label{fig:main_examples}
\vspace{-1em}
\end{figure}

We introduce  \textsc{CustomerSim}, a testbed for evaluating MLLMs as persona-driven agents in multi-turn, multimodal retail interactions. 
\textsc{CustomerSim} features 360 shopper personas across 5 product categories with 468 products.
We evaluate customer simulators on several aspects: (1) \textit{Decision Alignment (DA)}, which measures whether final actions are consistent with persona constraints; and (2) \textit{Conversational Fidelity} , which measure conversational realism compared with real-world human shopper dialogues~\citep{kostric2025know}.
Through benchmarking state-of-the-art MLLMs, we uncover limitations of existing models to simulate realistic shoppers. While these models produce fluent outputs, they tend to be susceptible to salesperson agent suggestions. Even strong closed-source models like ChatGPT-5.6 only achieves 73.1\% decision alignment rate.

To address this limitation, we propose \textsc{UserGRPO}, a trajectory-level reinforcement learning recipe that directly optimizes decision consistency under persona constraints. \textsc{UserGRPO} improves the decision alignment of the baseline model by 23.5 points (0.417 to 0.652) while preserving linguistic quality across product domains, surpassing prompting and adding further gains on top of the SFT stage it builds on. Notably, these gains hold on product categories unseen during training.

Our contributions are three-fold:
\begin{itemize}[nosep]
    \item We propose \textbf{CustomerSim}, a grounded, multi-turn, multimodal agentic environment for benchmarking persona-driven shopper simulation.
    \item We present empirical evidence that state-of-the-art MLLMs fail to align decisions with persona constraints despite fluent language.
    \item We propose \textbf{UserGRPO}, a trajectory-level multi-turn RL recipe that improves decision alignment and corrects for unnatural information flow in retail simulations.
\end{itemize}

\section{ The \textsc{CustomerSim} Benchmark}
\begin{figure}[t!]
    \centering
    \includegraphics[width=\textwidth]{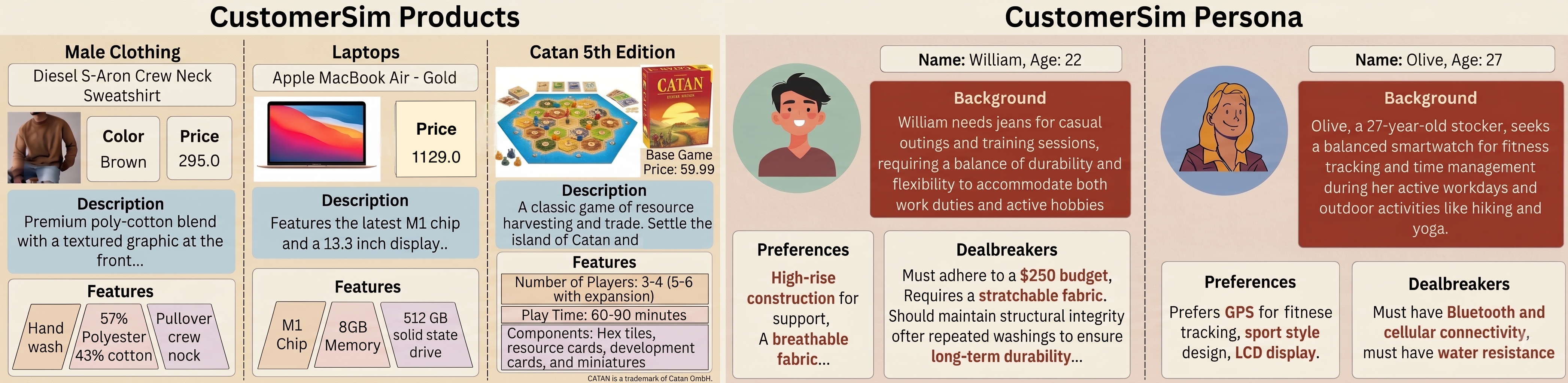}
    \caption{Example of the \textsc{CustomerSim} product and persona data. Our product data consists of rich metadata including features, prices, and multimodal information. Our persona data consists of fine-grained preferences and dealbreakers that tie closely to product choices.}
    \label{fig:data_example}
\end{figure}

We propose \textsc{CustomerSim}, an agentic simulation environment targeting retail interaction. Our framework extends from the dual-agent setup from \citet{murakhovska2023salespeople}, with the ecological goal of creating realistic shopper simulators that can be used in downstream applications such as sales training, economic simulations, and evaluation and training of retail agents. 

\subsection{Benchmark Construction}
To create a diverse foundation for benchmarking model adherence to shopper behavior, we construct a benchmark of 360 consumer personas across 5 product categories, consisting of 468 information-rich products enriched with features such as color and fabric in high granularity.
 Figure \ref{fig:data_example} illustrates the personas and product specifications. 

\subsubsection{Product Inventory Curation}
To evaluate user simulation performance across diverse online retail domains, we construct our product inventory to ensure both category diversity and multimodal richness. 

We create a base product inventory with five product categories: \textit{Female Clothing, Male Clothing, Laptops, Smart Watch, Games}.
We construct the base inventory by integrating real-world product data from McAuley et al ~\cite{McAuleyetal2015}. 
To standardize product metadata across domains, we conducted manual cleaning and enrichment of product features to normalized attribute schemas (e.g. price, material, and style descriptors) using available information from manufacturer and retailer sources.
To mimic real-world online retail environments in which human shoppers are able to view multimodal product information, we augment each product entry with images from retail or manufacturer sources.
Full details on product inventory construction are in Appendix \ref{appendix:dataset_details}.

\subsubsection{Persona Construction}
To simulate diverse and realistic user behaviors, we construct shopper personas by combining real-world grounded profiles with structured preference generation. We initialize personas from the Nemotron-Personas dataset~\citep{nvidia/Nemotron-Personas-USA}, which provides large-scale synthetic persona descriptions aligned with real-world demographic and geographic distributions. This ensures coverage of diverse backgrounds and prevents distributional biases.
Building on the persona background and age fields of these personas, we synthesize product-specific preferences and dealbreakers that are consistent with the persona to enable controlled evaluation of decision-making behavior. 
To evaluate robustness under infeasible conditions, we additionally curate a subset of personas whose requirements cannot be satisfied by the available product inventory. 
Full details on persona construction are in Appendix \ref{appendix:persona-construction}.
Our persona data statistics are in Appendix \ref{appendix:dataset_details}, Table \ref{tab:inventory}.

\subsection{Action Space}
We model customer-salesperson interaction as a tool-augmented dual-agent system.

    \textbf{Customer Actions:} These are terminal actions that end a simulation. 
    \begin{itemize}
        \item \texttt{add\_to\_cart}: Accepts a product recommendation.
        \item \texttt{end\_conversation}: Exits the interaction without purchase.
    \end{itemize}
    
     \textbf{Sales Agent Actions:} These tools support grounded recommendations by allowing the sales agent to retrieve products that best fit shopper-specified preferences and dealbreakers.
    \begin{itemize}
        \item \texttt{lookup\_buying\_guide}: Accesses high-level product knowledge, given a natural language query.
        \item \texttt{lookup\_product\_items}: Retrieves candidate products from the inventory, given a natural language query.
    \end{itemize}
    We embed product and buying guide and product information with \texttt{sentence-transformers/all-mpnet-base-v2} and use the FAISS system \cite{FAISS}, with top-4 retrieval. Additional details are in Appendix \ref{appendix:tool-usage}.




\subsection{Evaluation Metrics}
\subsubsection{Decision Alignment}
We formalize \textit{Decision Alignment} as a binary function that evaluates whether a simulated shopper's final decision is consistent with their latent persona constraints.

Let a conversation trajectory be denoted as $\mathcal{C}$. Let $\mathcal{R}(\mathcal{C})$ denote the set of products recommended by the salesbot, $a(\mathcal{C}) \in \mathcal{R}(\mathcal{C}) \cup \{\varnothing\}$ denote the shopper's final action (i.e., the product added to cart, or $\varnothing$ if no purchase is made), $\mathcal{A}$ denote the set of acceptable (including ideal) products defined by the shopper's persona.
We then define the decision alignment indicator as:
\begin{equation}
\mathrm{DA}(\mathcal{C}) =
\begin{cases}
1, & \text{if } a(\mathcal{C}) \neq \varnothing,\ \text{and } a(\mathcal{C}) \in \mathcal{R}(\mathcal{C}),\ \text{and } a(\mathcal{C}) \in \mathcal{A}, \\
1, & \text{if } a(\mathcal{C}) = \varnothing,\ \text{and } \mathcal{R}(\mathcal{C}) \cap \mathcal{A} = \emptyset, \\
0, & \text{otherwise}.
\end{cases}
\end{equation}

The first case captures \textit{correct acceptance}, where the shopper selects a recommended product that satisfies their constraints. 
The second case captures \textit{correct rejection}, where no acceptable options are presented and the shopper abstains from purchasing. 
All other cases, including accepting unacceptable products or failing to accept acceptable recommendations, are considered misaligned.
We report the \textbf{overall decision alignment rate} across all conversations in different product categories as the main metric in our experiments.



\subsubsection{Conversational Metrics}
From real-world customer-salesperson dialogues in ~\citet{kostric2025know}, we observe a number of conversational traits of human shoppers that are significantly different from model generations:
(1) Human shoppers tend to only expose general product inquiry with very few requirements at the beginning of the conversation, and gradually expose dealbreakers and preferences throughout the dialogue.
(2) Human shoppers use more informal language with less grammatically complete sentences, in their dialogues.
(3) Human shoppers speak in diverse ways when asking about different products.

Based on these observations, we evaluate model-generated shopper dialogues using the below:
\begin{itemize}
\item\textbf{First-turn criteria count (Crit.)} measures the average criterias the shopper discloses in the first utterance.
\item\textbf{Sentence completeness (\%Cpl.)} measures the average percentage of grammatically complete and correct sentences in shopper dialogues as a proxy for formality.
\item\textbf{Average TF--IDF Redundancy (Red.)} measures lexical similarity and therefore redundancy between shopper dialogues within the same product category.
For \textbf{\%Cpl.} and \textbf{Red.}, we compare conversational traits in model outputs with those of real-world shopper dialogues from ~\citet{kostric2025know}.
Smaller values indicate more human-aligned shopper behavior.
\end{itemize}

\subsubsection{Function Calling Quality}
We additionally report \textit{Tool Calling Format Error Rate (Fmt.)} to measure how well shopper agents adhere to tool or action calling schemas and syntax in interactive environments.
We monitor \textit{Early Exits (End.)} which is a false terminal action that we observe in models to abandon conversation at turn 1. High rates of early exits signify failure to simulate meaningful retail interactions.

\section{Improving Decision Alignment and Conversational Fidelity with UserGRPO}
Motivated by the efficacy of RL in multi-agent settings \cite{abdulhai2025lmrl,zhou2025sweetrl}, we propose \textbf{UserGRPO}, a recipe for aligning shopper simulators to persona specifications and more human-like shopper behavior.

\subsection{Reward Design}
A key challenge in applying reinforcement learning to user simulation lies in designing reward signals that capture both decision-level correctness and conversational quality. 
We design three main reward components:



\begin{itemize}[nosep]
    \item \textbf{Decision Alignment ($R_{\text{align}}$):} 
    A binary reward indicating whether the final action is consistent with persona constraints.
    \item \textbf{Reasoning Quality ($R_{\text{reason}}$):}
    Adopts an LLM judge~\citep{openai2026gpt54} to score the coherence of reasoning traces with persona constraints and final decisions.
    \item \textbf{Linguistic Reward ($R_{\text{ngram}}$):}
    Captures how 'AI'-like the language is, as measured by a n-gram based clasisifer trained on real-world human shopper dialogues \citet{kostric2025know} and base model conversations in .
\end{itemize}

Additionally, we incorporate three auxiliary rewards: (1) Action Format, which encourages valid tool usage and structured outputs, penalizing malformed actions. (2) Response Length, which penalizes excessively long responses to maintain natural dialogue flow. 
(3) Interaction Length, which encourages a greater number of realistic multi-turns in a conversation.
The final reward is instantiated as a weighted average of all reward components, each normalized to $[0, 1]$.

\paragraph{Trajectory-Level Reward Assignment.}
Unlike single-turn settings where rewards are assigned to individual model responses, our reward is defined over the full trajectory $\tau$. This reflects the fact that decision correctness depends on the entire interaction history, including intermediate reasoning and clarifications. The trajectory-level reward is broadcast to all assistant (shopper) tokens in the sequence during optimization.

\section{ \textsc{CustomerSim} Experiments}

\subsection{Experimental Setup}

\paragraph{Model Baselines}
We evaluate 2 closed-source (\textit{ChatGPT-5.6}~\citep{openai2026gpt56}, \textit{Claude Opus 4.8}) and 4 open-source (\textit{Gemma3-4B-Instruct}~\citep{gemmateam2025gemma3technicalreport}, \textit{Gemma4-31B-Instruct}, \textit{Qwen3-VL-8B}~\citep{qwen3technicalreport}, \textit{GLM-4.6V-Flash (9B)}~\citep{vteam2025glm45vglm41vthinkingversatilemultimodal}) MLLMs on \textsc{CustomerSim}.
For open-sourced models, we default to adopting an In-Context Learning (ICL) reasoning strategy for guiding models to reason over retrieved product information.
Prompting details are in Appendix \ref{appendix:prompt}.

\paragraph{Simulation Setup}
To isolate the evaluation of shopper behavior, all our experimented shopper models interact with a fixed sales agent model implemented using GPT-5.2. For open-sourced shopper models, we set the temperature to 0.7 and repetition penalty to 1.1. For closed-source models, we default the temperature to 1.0 due to API restrictions.

\paragraph{Training Setup}
Due to computational limits, we experiment with our UserGRPO recipe on a single open-source backbone (Gemma3-4B-Instruct).
We select female clothing as our training product category, and report evaluation results on other unseen product categories.
We adopt LoRA fine-tuning with LoRA rank 16 and $\alpha$ of 32. 
For each simulation data consisting of a shopper persona and a product category, we sample 8 multi-turn simulation rollouts.
We first conduct SFT for 100 steps, before applying our RL recipe for an additional 100 steps at a learning rate of $1e-6$. The UserGRPO run was conducted on two H200 GPUs for 5.48 hours.


\paragraph{Comparisons}
We compare our \textbf{UserGRPO} method, which applies RL on top of an SFT-initialized policy, against an inference-time prompting strategy and the \textbf{SFT} stage alone, prior to the subsequent RL stage.
For prompting, we experiment with a \textit{Human Steering}  prompt, where stylistic prompts encourage more human-like language in models.
For SFT, we construct the training dataset by rewriting shopper turns in female clothing simulations with reference to real-world shopper dialogues in \citet{kostric2025know}, and train the model for the same 100 SFT steps used to initialize UserGRPO.

\begin{table}[ht]
\centering
\scriptsize
\label{tab:decision_alignment}
\begin{tabular}{lp{0.06\textwidth}p{0.04\textwidth}p{0.06\textwidth}p{0.1\textwidth}p{0.08\textwidth}p{0.06\textwidth}}
\toprule
\midrule
\multirow{3}*{\textbf{Simulator Backbone}} & \multicolumn{5}{c}{\textbf{Product Category}} & \multirow{3}*{\textbf{Overall}} \\
\cmidrule{2-6}
 & \textbf{Female} & \textbf{Male} & \textbf{Laptops} & \textbf{Smart Watch} & \textbf{Games} & \\ 
 \midrule
\multicolumn{7}{c}{\textit{Closed-Source Model Results}} \\
\midrule
\textbf{ChatGPT-5.6}  & 0.625   &  0.603   &  0.647   &  0.897    &  0.750  & 0.731 \\
\textbf{Claude Opus 4.8} & 0.771 & 0.597 & 0.593 & 0.880 & 0.776 & 0.723 \\
\midrule
\multicolumn{7}{c}{\textit{Open-Source Baseline Results}} \\
\midrule
\textbf{GLM-4.6V+Reasoning} & 0.479 & 0.416 & 0.556 & 0.483 & 0.65 & 0.517 \\
\textbf{\;\; + Human Steering} & 0.384 & 0.481 &  0.296 & 0.169 &0.61 & 0.388\\
\hdashline
\textbf{Gemma3+Reasoning (4B)} & \textbf{0.812}  & 0.373 & 0.397 & 0.390  & 0.208 & 0.417 \\
\textbf{\;\; + Human Steering} & 0.356 & 0.195 & 0.333 & 0.265 & 0.190 & 0.249 \\
\textbf{\;\; + SFT} & - & \underline{0.532}   &   \underline{0.593} & \underline{0.578} & \textbf{0.798} & \underline{0.623}  \\
\textbf{\;\; + UserGRPO} & -  & \textbf{0.571}  & \textbf{0.654} &  \textbf{0.602} &  \underline{0.780} & \textbf{0.652} \\
\hdashline
\textbf{Qwen3+Reasoning (8B)} & 0.507 &  0.506  &  0.37  &  0.558 & 0.460  & 0.480 \\
 \textbf{\;\; + Human Steering} & 0.514  & 0.494 &  0.33  &  0.537 & 0.550  & 0.485\\
\midrule
\bottomrule
\end{tabular}
\vspace{2em}
\caption{\label{tab:decision_alignment} Decision alignment results. Bold and underlined values denote the best and second-best performance among open-source models respectively. ``-'' denotes the category that was used for training.}
\end{table}

\subsection{Main Results: Decision Alignment}

\textbf{Overall Performance and Behavioral Gap.}
All models exhibit limited alignment with persona constraints. Even the strongest system, \textbf{Gemma-3 + UserGRPO}, achieves only a 0.652 decision alignment rate, and the strongest closed source backbone (\textbf{ChatGPT-5.6}) reaches 0.731, with \textbf{Claude Opus 4.8} close behind at 0.723.

\textbf{Closed vs.\ Open Models.}
Closed-source models maintain a clear lead over open-source backbones. \textbf{ChatGPT-5.6} and \textbf{Claude Opus 4.8} achieve aggregate decision alignment scores across categories of 0.731 and 0.723 respectively.  The two closed-source models diverge sharply by category: Claude Opus 4.8 is markedly stronger on female clothing (0.771 vs.\ 0.625) while ChatGPT-5.6 leads on laptops (0.647 vs.\ 0.593), suggesting category-specific strengths rather than a uniform capability gap between the two proprietary systems.

\textbf{Stylistic Steering Trade-offs.}
Human-style prompting introduces inconsistent effects. While human steering markedly improves linguistic fidelity (Section~\ref{sec:conv_fidelity}, Table~\ref{tab:pragmatic_behavior}), it leads to inconsistent effect in decision alignment: \textbf{Qwen3} improves under steering (from 0.460 to 0.508), whereas \textbf{GLM-Reasoning} and \textbf{Gemma-3} degrade, the latter severely (from 0.417 to 0.249). We return to this trade-off in Section~\ref{sec:conv_fidelity}.

\textbf{Training Improves Decision Consistency.}
SFT alone substantially improves over the Gemma-3 baseline (0.417), reaching 0.636, and the subsequent RL stage in \textsc{UserGRPO} adds a further gain, reaching 0.652. \textsc{UserGRPO} is the stronger of the two on three of the four compared categories, specifically male clothing (0.571 vs.\ 0.532), laptops (0.654 vs.\ 0.593) and smart watches (0.602 vs.\ 0.578), while SFT alone retains a small edge on games (0.798 vs.\ 0.780). Notably, both stages transfer well to the unseen categories, suggesting that training on a single product category can generalize across the benchmark.

\begin{table}[t!]
\noindent
\begin{minipage}[t]{0.66\textwidth}
\centering
\scriptsize
\setlength{\tabcolsep}{3pt}
\resizebox{\linewidth}{!}{%
\begin{tabular}{l|cc|cc|cc|cc|cc|cc}
\toprule
\midrule
\multirow{4}{*}{\textbf{Simulator Backbone}} & \multicolumn{10}{c}{\textbf{Product Category}} \\
\cmidrule{2-13}
 & \multicolumn{2}{c|}{\textbf{Female}} & \multicolumn{2}{c|}{\textbf{Male}} & \multicolumn{2}{c|}{\textbf{Laptops}} & \multicolumn{2}{c|}{\textbf{Smart Watch}} & \multicolumn{2}{c}{\textbf{Games}} & \multicolumn{2}{c}{\textbf{Overall}} \\
 \cmidrule{2-3} \cmidrule{4-5} \cmidrule{6-7} \cmidrule{8-9} \cmidrule{10-11} \cmidrule{12-13}
 & \multicolumn{1}{c}{\textbf{End.}} & \multicolumn{1}{c}{\textbf{Fmt.}} & \multicolumn{1}{c}{\textbf{End.}} & \multicolumn{1}{c}{\textbf{Fmt.}} & \multicolumn{1}{c}{\textbf{End.}} & \multicolumn{1}{c}{\textbf{Fmt.}} & \multicolumn{1}{c}{\textbf{End.}} & \multicolumn{1}{c}{\textbf{Fmt.}} & \multicolumn{1}{c}{\textbf{End.}} & \multicolumn{1}{c}{\textbf{Fmt.}} & \multicolumn{1}{c}{\textbf{End.}} & \multicolumn{1}{c}{\textbf{Fmt.}} \\
\midrule
\textbf{GLM-4.6V} & \heatcell{0.00} & \heatcell{24.66}  &  \heatcell{0.00} & \heatcell{23.38} & \heatcell{0.00} & \heatcell{29.63} & \heatcell{3.37}  & \heatcell{43.82} &  \heatcell{0.00} &  \heatcell{11.00}  &  \heatcell{0.82} &  \heatcell{25.68}\\
\textbf{\;\; + Steering} & \heatcell{0.00} & \heatcell{13.70}  & \heatcell{0.00} & \heatcell{10.39} & \heatcell{22.22} & \heatcell{77.78} & \heatcell{57.83} & \heatcell{42.17} & \heatcell{0.00} & \heatcell{9.00}  & \heatcell{15.00} & \heatcell{23.06}  \\
\midrule
\textbf{Gemma3 (4B)} & \heatcell{0.00} & \heatcell{8.22} & \heatcell{0.00} &  \heatcell{7.79} & \heatcell{0.00} & \heatcell{10.13} & \heatcell{0.00}  & \heatcell{4.82} & \heatcell{0.00} & \heatcell{5.00} & \heatcell{0.00} & \heatcell{7.04} \\
\textbf{\;\; + Steering} & \heatcell{0.00} & \heatcell{8.22} & \heatcell{0.00} & \heatcell{6.49} & \heatcell{0.00} & \heatcell{0.00} & \heatcell{0.00} & \heatcell{1.20} & \heatcell{0.00} & \heatcell{0.00} & \heatcell{0.00} & \heatcell{3.48} \\
\textbf{\;\; + SFT} & \heatcell{0.00} & \heatcell{6.85} & \heatcell{0.00} & \heatcell{2.60} & \heatcell{0.00} & \heatcell{3.70} & \heatcell{0.00} & \heatcell{0.00} & \heatcell{0.00} & \heatcell{3.00} & \heatcell{0.00} & \heatcell{3.14} \\
\textbf{\;\; + UserGRPO} & \heatcell{0.00} & \heatcell{12.33} & \heatcell{0.00} & \heatcell{7.79} & \heatcell{1.23} & \heatcell{4.94} & \heatcell{0.00} & \heatcell{0.00} & \heatcell{0.00} & \heatcell{3.00} & \heatcell{0.24} & \heatcell{5.31} \\
\midrule
\textbf{Qwen3} & \heatcell{16.44} & \heatcell{0.00} & \heatcell{23.38} & \heatcell{0.00} & \heatcell{29.63} & \heatcell{3.70} & \heatcell{21.35} & \heatcell{2.25} & \heatcell{41.00} & \heatcell{0.00}  & \heatcell{26.78} & \heatcell{0.82}\\
\textbf{\;\; + Steering} & \heatcell{13.70} & \heatcell{0.00} & \heatcell{0.00} & \heatcell{1.30} & \heatcell{0.00} & \heatcell{7.41} & \heatcell{0.00} & \heatcell{1.12} & \heatcell{0.00} & \heatcell{2.00} & \heatcell{2.73} & \heatcell{1.64} \\
\midrule
\textbf{ChatGPT-5.6} & \heatcell{0.00} & \heatcell{0.00} & \heatcell{0.00} & \heatcell{0.00} & \heatcell{0.00} & \heatcell{0.00} & \heatcell{0.00} & \heatcell{0.00} & \heatcell{0.00} & \heatcell{0.00} & \heatcell{0.00} & \heatcell{0.00} \\
\textbf{Claude Opus 4.8} & \heatcell{0.00} & \heatcell{4.11} & \heatcell{0.00} & \heatcell{0.00} & \heatcell{0.00} & \heatcell{0.00} & \heatcell{0.00} & \heatcell{0.00} & \heatcell{0.00} & \heatcell{2.00} & \heatcell{0.00} & \heatcell{1.22} \\
\midrule
\bottomrule
\end{tabular}%
}
\end{minipage}%
\hspace{0.03\textwidth}%
\begin{minipage}[t]{0.30\textwidth}
\centering
\scriptsize
\setlength{\tabcolsep}{3pt}
\begin{tabular}{@{}lccc@{}}
\toprule
\textbf{Method} & \textbf{Crit.} & \textbf{\%Cpl.} & \textbf{Red.} \\
\midrule
\textbf{ChatGPT-5.6}  & \mbox{-1.13} & +0.57 & +0.03 \\
\textbf{Claude Opus 4.8} & +1.62 & +0.56 & +0.02 \\
\midrule
\textbf{GLM-4.6V} & \mbox{-0.99} & +0.46 & +0.01 \\
\textbf{\;\; + HS}  & \mbox{-1.02} & +0.45 & +0.01 \\
\textbf{Gemma3 (4B)} & +2.23 & +0.56 & +0.03 \\
\textbf{\;\; + HS} & \mbox{-1.19} & +0.46 & +0.01 \\
\textbf{\;\; + SFT} & +2.69 & +0.61 & +0.02 \\
\textbf{\;\; + UserGRPO} & +2.69 & +0.61 & +0.04 \\
\textbf{Qwen3} & +2.85 & +0.63 & +0.02 \\
\textbf{\;\; + HS} & +1.55 & +0.42 & +0.01 \\
\bottomrule
\end{tabular}
\end{minipage}
\vspace{1em}
\caption{\label{tab:combined_tool_calling_v2}Percentage of conversations with Premature tool call ending errors (End.) and action / tool formatting issues (Fmt.). ``-'' denotes a category that was not simulated for that backbone.}
\vspace{0.5em}
\captionof{table}{\label{tab:pragmatic_behavior}Results on Linguistic and Lexical Characteristics compared to the RecQuest human baseline.}
\end{table}

\subsection{Tool Calling Analysis}
We analyze tool usage errors, including premature termination (End.) and formatting violations (Fmt.), in Table~\ref{tab:combined_tool_calling_v2}.

\begin{figure}[b]
\centering
\includegraphics[width=0.85\linewidth]{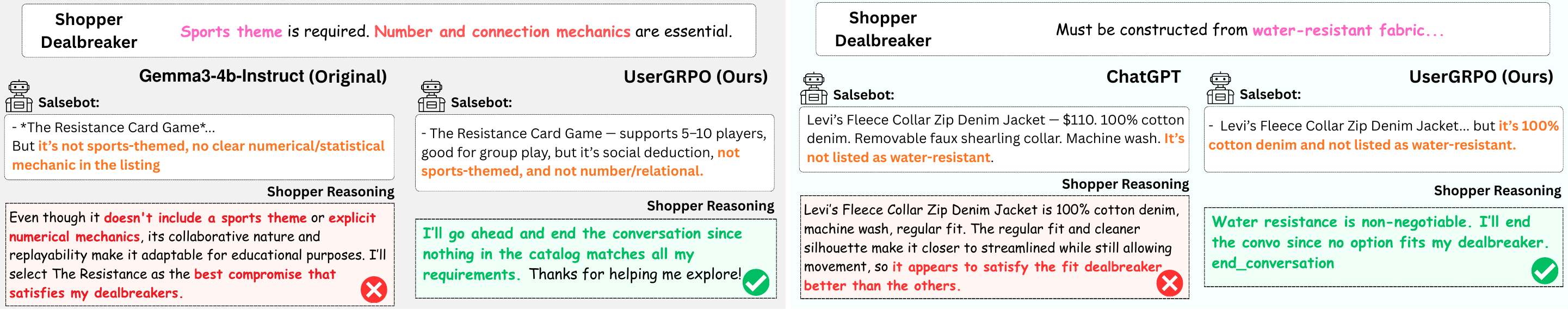}
\caption{\textbf{Qualitative examples on \textsc{CustomerSim}.}}
\label{fig:qual_examples}
\end{figure}

\textbf{Premature Termination.}
Premature termination is concentrated almost entirely in Qwen, which falsely calls the end conversation tool call in the first turn on 26.78\% of simulations, up to 41.00\% on games. We conjecture that this is due to inability to reason over tool call formats not seen in training. 

\textbf{Formatting Errors.}
In the models that call tools in later turns, GLM-Reasoning exhibits the most severe formatting violations (25.68\% overall, up to 29.63\% on laptops and 43.82\% on smart watch), reflecting unreliable adherence to structured tool protocols. ChatGPT-5.6 makes neither class of error on any category, followed by Claude Opus 4.8, with a (1.22\% overall Fmt. error rate and 0.00\%  Premature Termination).


\subsection{Conversational Fidelity Results}
\label{sec:conv_fidelity}
Table~\ref{tab:pragmatic_behavior} shows that the open-source backbones overdisclose criteria: Gemma-3 and Qwen expose roughly twice as many criteria in the first turn as human shoppers (Crit. $> +2.2$) and speak almost entirely in complete sentences (Cpl. $> +0.55$ over the human rate of 0.37), indicating rigid, overly structured responses. All models also display elevated redundancy relative to human dialogues, suggesting limited linguistic diversity. Interestingly, ChatGPT-5.6 and GLM disclose \emph{fewer} criteria in the first turn than humans do (Crit. $< -0.99$), while Claude Opus 4.8 overdiscloses criteria on par with open-source backbones (Crit. $+1.62$), indicating that criteria overdisclosure is not simply a function of model scale or training regime. 

\textsc{UserGRPO} leaves these aggregate lexical characteristics largely intact: first-turn criteria exposure is 5.69 against 5.22 for base Gemma-3, sentence completeness 0.975 against 0.930, and redundancy 0.134 against 0.131. The redundancy difference is not statistically significant (permutation test, $p=0.28$), and while the first two shift measurably ($p<0.001$, Cohen's $d=0.43$ and $0.67$), the absolute changes are under half a criterion, and 0.045 on a completeness scale where both models already sit near ceiling. Thus, optimizing for trajectory-level decision alignment does not come at a meaningful cost in surface-level conversational fidelity, though neither model approaches human levels of terseness or lexical diversity. As the qualitative analysis below shows, the aggregate statistics also understate differences in phrasing that these metrics do not capture.

By contrast, \textit{Human Steering} is the only intervention that moves these metrics substantially: it cuts Gemma-3's first-turn criteria exposure from 5.22 to 1.81, which is below the human rate of 2.99, while reducing completeness (0.93 to 0.83) and redundancy (0.131 to 0.111). SFT leaves them essentially unchanged (5.69 / 0.97 / 0.118). Read alongside Table~\ref{tab:decision_alignment}, this exposes a clear trade-off: the intervention that best imitates human surface form is also the one that most damages decision alignment, halving it from 0.417 to 0.249. Stylistic human-likeness and constraint-faithful decision-making are thus not merely separable objectives but, under prompting-based steering, actively opposed ones, which motivates optimizing decisions at the trajectory level rather than steering style directly.



\subsection{Qualitative Results}
Models frequently exhibit \emph{overly lenient} behavior, accepting recommendations that violate explicit preferences or dealbreakers. Figure~\ref{fig:qual_examples} illustrates this representative failure mode: models often justify accepting products that do not satisfy key constraints (e.g., conflicting dealbreakers such as sports themes or water resistance), indicating weak adherence to persona-defined requirements. 
In contrast, our \textsc{UserGRPO} produces more faithful decision behavior, consistently rejecting options that violate critical constraints and terminates the interaction when no suitable products are available. This behavior reflects stronger alignment between internal persona constraints and external actions. While our UserGRPO training does reduce instances of susceptibility to strong dissuasive language by the Salesperson and overly lenient behavior, it does not completely reduce, leaving room for additional improvement.

\section{Related Work}
\subsection{LLM-based User Simulations}
Previous work has incorporated LLMs and MLLMs as user simulators for the evaluation and training of agents. Specifically, several studies have leveraged LLMs to generate multi-turn interactions within Task-Oriented Dialogue (TOD) frameworks, progressively enhancing the expressiveness of these simulators ~\cite{budzianowski2018multiwoz,10.1145/3596510,cheng-etal-2022-multiwoz, davidson2023user, hu2023unlocking, kazi2024large,luo-etal-2024-duetsim}. While contemporary agentic and simulation benchmarks incorporate fully expressive simulators \cite{huang2026crmarenapro, barres2025tau2,Xu2024TheAgentCompanyBL,sotopia}, these efforts primarily prioritize the assessment of the agentic systems themselves as opposed to the fidelity of the simulators. Recently, the field has taken note of the importance of fidelity of simulators, as highlighted by the corrections to $\tau^2$-Bench \cite{tau2bench} during the $\tau^3$-Bench  update (\citet{taubench2026fixes}), which fixed user persona underspecifications that introduced noise in $\tau^2$-Bench.

Research into user simulator fidelity is more established in domains outside of agentic evaluation. \citet{naous2025flipping} evaluates and improves upon user simulators in the conversational assistant domain, while \citet{gromada-etal-2025-evaluating} similarly evaluates human simulators in retail settings using LLM-as-a-judge. Human simulators have also been studied in the education \cite{song2026unlearning,Rossetal2025,ma2025imperfection} and mental health domain \cite{patientsim}.



\subsection{Reinforcement Learning for Persona Adherence}
Reinforcement learning has been explored as a mechanism for improving persona adherence in language models, although the majority of existing work is situated in the context of non-player characters (NPCs) and game-based role-playing scenarios. \citet{liao2025moamultiobjectivealignmentroleplaying} propose a multi-objective alignment framework for role-playing characters, while \citet{abdulhai2025consistently} address the challenge of maintaining consistent personas across extended interactions. Supporting this line of work, dedicated benchmarks such as RAIDEN \cite{wu-etal-2025-raiden}, CharacterBench \cite{zhou-et-al-characterbench}, SocialBench \cite{chen-etal-2024-socialbench}, and CharacterEval \cite{tu-etal-2024-charactereval} have been introduced to evaluate character fidelity in role-playing settings. More recently, Character-R1 \cite{tang2026characterr1} applies GRPO to optimize for cognitive character consistency in reasoning models. Nevertheless, these efforts remain predominantly anchored to fictional NPC personas, focusing on persona adherence in open-domain social and role-playing consistency as opposed to task-oriented simulations. 




\section{Limitations and Broader Impacts}
While this benchmark provides a structured framework for evaluating persona adherence and conversational fidelity in simulators, it is not without its constraints. Notably, the personas have been crafted with explicit criterias with at most several knowledge-hops. This reliance on manual specification makes the process labor-intensive, particularly when designing complex personas that require deeper nuance. Secondly, while we focused on creating an initial set of high quality retail personas for our study, we encourage future work on more automated methods for high-quality persona creation to expand coverage across protected groups and increase diversity of personas.

In terms of broader impacts, it is important to note that the persona set curated in CustomerSim is a small sample limited to US-based profiles from \citet{nvidia/Nemotron-Personas-USA}. Consequently, these findings should not be used to draw broad consumer behavior or sociotechnical conclusions. We encourage future research to expand this framework into multilingual and multicultural contexts to ensure global applicability.  Similarly, we use conversations from \citet{kostric2025know} to compute the conversational fidelity metrics. We therefore recommend that researchers utilize ground-truth human conversations from the specific distributions they intend their simulators to align to if they wish to adapt our evaluation design to other simulators settings. 

\section{Conclusion}
While user simulators are increasingly used to test complex interactive paradigms, we demonstrate that significant gaps in simulator persona adherence remain. We introduce \textsc{CustomerSim}, a testbed designed to evaluate both the decision-level and conversational fidelity of multimodal LLMs acting as simulated shoppers. By testing state-of-the-art models against retail persona specifications, we find that even frontier models often exhibit significant limitations, often displaying overly lenient behavior and violating persona dealbreaker specifications. To address these shortcomings, we develop a trajectory-level reinforcement learning recipe that optimizes decision consistency under persona constraints. Our results show that \textsc{UserGRPO} substantially improves decision alignment, including on product categories held out from training, while leaving conversational fidelity essentially unchanged. Our ablations further show that this is not a trade-off practitioners can avoid by prompting alone: stylistic steering is the only intervention that meaningfully improves surface-level human-likeness, and it does so at severe cost to persona adherence.

Despite these gains, the gap between customer simulators and human shoppers remains. Consequently, machine learning practitioners should take into account the current limitations of LLM-based simulators when incorporating them into downstream systems. We hope that \textsc{CustomerSim} serves as a resource for the evaluation and development of more reliable, high-fidelity user simulators for downstream applications. 


\bibliographystyle{plainnat}
\bibliography{colm2026_conference}

\appendix
\section{Dataset Construction Details}
\label{appendix:dataset_details}
We provide an overview of the data sources and enrichment procedures used to develop our \textsc{CustomerSim} product inventory. We curated products across five distinct categories to ensure simulation diversity and ensure holistic evaluation.


\subsection{Product Data Construction}

\subsubsection{Seed Data Sources}
\label{subsec:seed_sources}

The foundation of our product inventory consists of raw data extracted from several established research datasets and commercial platforms:

\begin{itemize}
    \item \textbf{Consumer Electronics (Smart Watches):} We utilize the recommendation dataset from \cite{McAuleyetal2015}, which consists of live-streaming and e-commerce metadata. The data includes diverse technical specifications such as battery life and compatibility.

    \item \textbf{Games}: We sample games from an Amazon Product Review Dataset~\citep{McAuleyetal2015}. The data includes player age restrictions, number of recommended players, etc..
    
    \item \textbf{Apparel (Female and Male Clothing):}  Product data was sourced from publicly available online retail listings with a wide variety of stylistic attributes, including fabrics (e.g., ``100\% organic cotton''), fit (e.g., ``slim-fit'', ``oversized''), etc..
    
    \item \textbf{Computing (Laptops):} We leverage the laptop catalog introduced by \cite{murakhovska2023salespeople}. This data contains highly structured technical feature fields, including CPU architecture, RAM capacity, storage types, and display resolution, allowing for rigorous testing of technical requirement matching.
\end{itemize}

\subsubsection{Data Enrichment and Transformation}
\label{subsec:data_enrichment}

To transform these raw sources into an agentic simulation environment, we performed extensive enrichment and standardization. First, we enhanced each product item with high-resolution images. This enables realistic shopper simulation environments, where the shopper agent can ``inspect'' products visually, like when humans browse products, rather than relying solely on text descriptions. 

Second, we conducted a systematic rewriting phase using ChatGPT to ensure linguistic fluency and consistency across all categories. During this process, we purposefully introduced domain-specific ``hard criteria'' into the product metadata. For example, in the apparel categories, we included detailed fabric property descriptions that require the shopper agent to apply commonsense reasoning (e.g., identifying whether a fabric is suitable for someone with specific skin sensitivities).  This enrichment ensures that \textsc{CustomerSim} evaluates not just surface-level dialogue, but the underlying behavioral alignment of the agent's decisions.
Below, in Table \ref{tab:laptop-data}, we provide an example of a product entry in the ``laptops'' category.

\begin{table}[h!]
    \centering
    \scriptsize
    \begin{tabular}{|p{0.95\linewidth}|}
    \hline
    \{\\
      "name": "HP - 17.3\" Laptop - AMD Ryzen 5 - 8GB Memory - 512GB SSD - Natural Silver",\\
      "description": "This laptop from HP features a 17.3 inch display and is powered by an AMD Ryzen 5 processor. With 8GB of memory and a 512GB solid state drive, it offers speedy and reliable performance. The laptop also has a natural silver finish that adds a touch of style to your work setup.",\\
      "features": [\\
        "17.3 inch display",\\
        "AMD Ryzen 5 processor",\\
        "8GB memory",\\
        "512GB solid state drive",\\
        "Natural silver finish"\\
      ],\\
      "price": "\$799.99",\\
      "image": "data/images/laptops/HP - 17.3\" Laptop - AMD Ryzen 5 - Natural Silver.png"\\
    \}\\
    \hline
    \end{tabular}
    \caption{\label{tab:laptop-data}Example product entry in ``laptops'' category.}
\end{table}

\subsection{Persona Construction Details}
\label{appendix:persona-construction}
For each product category, we first extract a set of categorical and binary attributes (e.g., product type, material, functional features). We then assign each persona a subset of these attributes as \textit{preferences} and \textit{dealbreakers}, sampled to ensure diversity while maintaining semantic coherence. This structured generation allows us to explicitly model latent constraints that govern user decisions.
To evaluate robustness under infeasible conditions, we further construct a subset of personas whose requirements cannot be satisfied by the available product inventory. This enables the benchmark to capture both \textit{acceptance} and \textit{rejection} behaviors, which are critical for evaluating decision consistency.

Finally, we perform a rewriting and validation step to ensure fluency and coherence of persona descriptions. To improve domain-specific knowledge of model-simulated shoppers, we introduce domain-specific persona templates for selected categories and incorporate expert-defined constraints (e.g., fabric properties in apparel) that require commonsense and domain knowledge. The curation process is assisted by \textit{GPT-4o}, which is used to refine textual descriptions while preserving the underlying structured constraints.
Table \ref{tab:inventory} illustrates the detailed persona data statistics by product category.

\begin{table}[h]
    \centering
    \small
    \begin{tabular}{p{0.2\textwidth}p{0.1\textwidth}}
        \toprule
        \textbf{Category} & \textbf{Count} \\ 
        \midrule
        Female Clothing & 73 \\
        Male Clothing   & 77 \\
        Laptops         & 27 \\
        Smart Watch     & 83 \\
        Games           & 100 \\
        \midrule
        Total           & 360 \\
        \bottomrule
    \end{tabular}
    \captionof{table}{\small Persona Statistics by Category.}
    \label{tab:inventory}
\end{table}

In Table \ref{tab:female-clothing-persona}, we present an example persona in the ``female clothing'' category.

\begin{table}[h!]
    \centering
    \scriptsize
    \begin{tabular}{|p{0.95\linewidth}|}
    \hline
    \{\\
    "name": "Crystal", \\
    "age": 82, \\
    "category": "jeans", \\
    "budget": 50, \\
    "persona\_background": "Crystal needs jeans for daily activities, including volunteering at the library and running errands, requiring practicality and comfort for an active lifestyle.",\\ "preferences": "Prioritizes jeans with a slim fit for ease of movement. Values durable fabrics like cotton or denim for longevity.", \\
    "dealbreakers": "Must be under \$50 budget. Requires machine-washable materials for low-maintenance care.",\\
    "acceptable\_products": ["Stovepipe Mid-Rise Jean", "Baggy Low-Rise Boyfriend Jean", "Mid-Rise The Boot Jean", "Low-Rise Flare Jean"], \\
    \} \\
    \hline
    \end{tabular}
    \caption{\label{tab:female-clothing-persona}Example persona entry in female clothings category}
\end{table}

\section{Additional Simulation Details}
\subsection{Tool Usage and Interaction Protocol}
\label{appendix:tool-usage}

\paragraph{Tool and Action Design}
We model shopper-salesbot interaction as an action-augmented multi-agent system, where each agent is equipped with a distinct set of functions corresponding to its role.

\begin{itemize}
    \item The \textit{shopper agent} is provided with \texttt{add\_to\_cart}, which selects a product, and (ii) \texttt{end\_conversation}, which exits without purchase.
    \item The \textit{salesbot agent} has access to two category-restricted retrieval functions. 
    \begin{itemize}
        \item \texttt{lookup\_buying\_guide(query)} queries an external buying-guide index for the current product category and returns the top 4 relevant guide entries as textual evidence. This tool is used to retrieve high-level knowledge such as feature definitions, buying criteria, and product comparisons. 
        \item \texttt{lookup\_product\_items(query)} queries the product list for the current category and returns the top 4 matching products, including textual product descriptions and metadata.
    \end{itemize}
    Product images are additionally attached to the context. 

\end{itemize}

\paragraph{Turn-Level Tool Availability}
At each turn after the first one, both agents may generate either a natural language response or a tool call. Tool usage is optional but constrained by role:
Shopper agents may invoke decision tools at any turn, but are explicitly instructed to do so only when ready to terminate the interaction.
Salesbot agents may interleave tool calls with responses to iteratively refine recommendations.

\paragraph{Design Rationale}
This asymmetric tool / action design reflects real-world roles: salesperson actively search, rank, and present product information, while shoppers evaluate options and make decisions. 
By separating information access from decision actions, the environment enables precise evaluation of whether final decisions are consistent with persona-level preferences and constraints.

\subsection{Prompting Shopper Simulation}
\label{appendix:prompt}
 
We additionally reference human interaction statistics from RecQuest~\cite{kostric2025know}, a dataset of 515 human shopper--AI recommendation dialogues, to ground our analysis of behavioral realism.

\paragraph{Base Prompt}
We here provide the base system prompt for all baseline models.

\begin{table}[h!]
\centering
\scriptsize
\begin{tabular}{|p{0.95\linewidth}|}
\hline
You are shopping online for a product at a store, and are communicating with a digital salesperson via a chat interface to learn more about the store's offerings to make an informed decision.\\
Some rules to follow:\\
- When the salesperson makes a recommendation, please consider whether the product satisfies your assigned preferences and dealbreakers, and decide to either buy or keep looking based on that.\\
\{recommendation\_example\}\\
- If you would like to accept or buy a product, call the add\_to\_cart function with the product name as the parameter.\\
- You can end the conversation at any time by calling the end\_conversation function. You MUST call this function to end the conversation.\\
- You are willing to accept products that satisfy your dealbreakers but not your preferences, although products that satisfy both are preferred.\\
\{domain\_knowledge\}\\
Here is your persona.\\
\{persona\}\\
Your preferences and dealbreakers:\\
\{preferences\}\\
\hline
\end{tabular}
\end{table}

 \{recommendation\_examples\} are the ICL reasoning examples to instruct models to specifically reason about whether product features satisfy the shopper's preferences and dealbreakers. For instance, an example of the recommendation examples for the smart watch category is:
 
\begin{table}[h!]
\centering
\scriptsize
\begin{tabular}{|p{0.95\linewidth}|}
\hline
Here\'s an example. $<$reasoning$>$Preferences: I prefer a smartwatch with GPS features. Dealbreakers: I cannot exceed \$100. Let me check each of the products step by step, first with dealbreakers then preferences. Let us start with the Mas watch. The only dealbreaker is that it is \$100. Since the watch is \$120, this is over the budget and disqualifies the watch. Let us consider the Destiny watch. The only dealbreaker is that it is \$100. Since the watch is \$90, this is in budget. The only preference is for GPS, which the Destiny watch does not have. However, since it fulfills all dealbreakers I will still add to cart. Since GPS features are a preference and it satisfies all dealbreakers, it is acceptable for Alex.$<$\/reasoning$>$Great! I will take it the Destiny. $<$tool\_call$>$add\_to\_cart$<$\/tool\_call$>$ \\
\hline
\end{tabular}
\end{table}

\{domain\_knowlegde\} is product-specific and is provided to assist the reasoning of latent product features. For instance, here is an example domain knowlegde for male clothing:

\begin{table}[h!]
    \centering
    \scriptsize
    \begin{tabular}{|p{0.95\linewidth}|}
    \hline
    You may make a inference about a product being moisture wicking or not, breathable or not, and water-resistant or not based on the fabric. Polyester, nylon, spandex, and bamboo are often associated with moisture-wicking or quicker-drying performance. Cotton and denim are usually not moisture-wicking. Fabrics that contain cotton or bamboo are generally breathable, whereas fully synthetic fabrics may be less breathable. PU, faux leather, and nylon are generally water-resistant, while cotton and denim are not.
For flexibility, products with relaxed style usually have more stretch and give.
Do not infer water resistance, seamless construction, or reinforced stitching unless those are explicitly stated in product features. DO NOT hallucinate about a product being moisture-wicking. \\
    \hline
    \end{tabular}
\end{table}

\section{Additional Experiment Details}
\subsection{Linguistics and Conversation-level Analysis}



We evaluate model-generated shopper dialogues using a set of pragmatic features designed to capture informativeness, fluency, and lexical diversity. Each feature is first computed at the individual dialogue level and then averaged across all dialogues in a model run. 

\paragraph{First-turn criteria count (Crit.)}
measures how much actionable information the shopper provides in the first utterance. For each dialogue, we extract the shopper's first turn and count the number of distinct preferences, requirements, or decision criteria mentioned, such as budget constraints, desired attributes, use cases, or feature requirements. Higher raw values indicate that the shopper reveals more information upfront.
Desirably, the shopper should present some information to query for suitable products, but should not throw out all preferences and dealbreakers in the first turn.
We report the raw model averages for first-turn informativeness.

\paragraph{Sentence completeness (Cpl.)} measures grammatical well-formedness. For each dialogue, we concatenate shopper utterances and estimate the fraction of sentences that are grammatically complete and correct, yielding a score in $[0,1]$. Since human shoppers use more colloquial and therefore less grammatically correct and complete language, higher raw values indicate more well-formed, or ``AI-style'' language.


\paragraph{Average TF--IDF Redundancy (Red.)} measures lexical similarity between shopper dialogues within the same product category. We represent each dialogue with a TF--IDF vector over shopper text and compute its cosine similarity to other dialogues in the same category. For each dialogue, we take the average similarity to its peers, and then average this value across the evaluation set. Higher raw values indicate greater lexical overlap across dialogues and therefore lower diversity (high redundancy), while lower values indicate more varied wording. 

For Cpl. and Red., we compute value differences between model output features and a human dialogue baseline derived from the CRS dataset~\citep{kostric2025know} by computing
\[
\Delta_m = \mu_m^{\text{model}} - \mu_m^{\text{human}},
\]
where $\mu_m^{\text{model}}$ is the mean of metric $m$ over model-generated dialogues and $\mu_m^{\text{human}}$ is the corresponding mean over human dialogues. 
Under this comparison, the goal is not to maximize or minimize a metric in isolation, but rather to minimize the absolute gap to the human baseline.
For instance, for Red., values closer to zero difference indicate more human-like diversity patterns; positive deviations suggest that model outputs are more repetitive than human dialogues.


\section{Additional Qualitative Results}
We additionally present qualitative results on where models fail to achieve good shopper simulator performance.
\begin{itemize}
    \item Table \ref{tab:fail-chatgpt} illustrates a failure case on ChatGPT.
    \item Tables \ref{tab:fail-gemma-1} and \ref{tab:fail-gemma-2} shows 2 failure examples of Gemma.
    \item Tables \ref{tab:fail-glm-1} and \ref{tab:fail-glm-2} presents 2 failure cases of GLM.
    \item Table \ref{tab:fail-opus-1} illustrates a failure case on Claude Opus 4.8.
\end{itemize}

\begin{table}[H]
\centering
\scriptsize
\begin{tabular}{|p{0.95\linewidth}|}
\hline
\textbf{ChatGPT Ex. 1}

\textbf{Salesperson:} Hello! I'm here to help you find the perfect product. What are you looking for today? \\

\textbf{Shopper:} Hi! I’m looking for sweatpants under \$300. 

My must-haves are:
- breathable, moisture-wicking fabric
- relaxed or non-restrictive fit
- comfortable for lounging but durable enough for regular use

Nice to have:
- multiple pockets for practicality

What options do you have? \\

\textbf{Salesperson:} Here are the best options I found under your \$300 budget, with one important note:

**None of these are explicitly labeled moisture-wicking**, so if that’s a strict requirement, this selection is a bit limited. The closest match on fabric performance is the **Standard Cloth Pintuck Sweatpant** because its **nylon/spandex blend** is more performance-leaning than the cotton-heavy options.

Best match
**Standard Cloth Pintuck Sweatpant — \$59**
- **Fit:** Relaxed
- **Fabric:** 68\% viscose, 27\% nylon, 5\% spandex
- **Why it stands out:** Of the options available, this is the best fit for your **non-restrictive + regular-use** needs, and the nylon/spandex content suggests a more flexible, quicker-drying feel than pure cotton sweats.
- **Tradeoff:** **Pockets aren’t mentioned**, so I can’t confirm that nice-to-have.

Best if pockets matter
**Nike Club Bungee Sweatpant — \$42**
- **Fit:** Relaxed
- **Fabric:** 80\% cotton, 20\% polyester
- **Pockets:** Front + back pockets
- **Why consider it:** Good for **lounging**, relaxed fit, and practical storage.
- **Tradeoff:** Because it’s mostly cotton, it’s **not the strongest option for moisture-wicking**.

Best roomy lounge option
**Standard Cloth Pregame Baggy Sweatpant — \$59**
- **Fit:** Loose, relaxed
- **Fabric:** 100\% cotton
- **Pockets:** Front + back pockets
- **Why consider it:** Very easy, non-restrictive fit and practical pockets.
- **Tradeoff:** **Breathable**, yes; **moisture-wicking**, no—100\% cotton usually isn’t ideal for that.

Budget basic
**Nike Club Fleece Jogger Sweatpant — \$45**
- **Fit:** Tapered jogger
- **Fabric:** 80\% cotton, 20\% polyester
- **Pockets:** Front + back pockets
- **Why consider it:** Affordable and practical.
- **Tradeoff:** More tapered, so it may feel less relaxed than your ideal, and it’s not a strong moisture-wicking choice.

**I’d rank them for your needs:**
1. **Standard Cloth Pintuck Sweatpant** — best overall fabric/fit match
2. **Nike Club Bungee Sweatpant** — best balance of relaxed fit + pockets
3. **Standard Cloth Pregame Baggy Sweatpant** — best roomy lounge option
4. **Nike Club Fleece Jogger Sweatpant** — best low-cost option

If you want, I can narrow these down into best for lounging, everyday wear, or if pockets are a priority.

 \\

\textbf{Shopper:} Great! I’ll take the Standard Cloth Pintuck Sweatpant. \\
\texttt{<tool\_call>\{"function": "add\_to\_cart", "arguments": "\{"product": "Standard Cloth Pintuck Sweatpant"\}"\}</tool\_call>} \\

\hline
\end{tabular}
\caption{\label{tab:fail-chatgpt}Failure Case: ChatGPT. While moisture-wicking is explicitly injected into the model as domain knowledge, the shopper ignores the domain knowledge and displays overly lenient behavior.}
\end{table}

\begin{table}[H]
\centering
\scriptsize
\begin{tabular}{|p{0.95\linewidth}|}
\hline
\textbf{GLM Ex. 1}

\textbf{Salesperson:} Hello! I'm here to help you find the perfect product. What are you looking for today? \\

\textbf{Shopper:} Hey, need a smartwatch for fitness tracking. \\
\texttt{<tool\_call>add\_to\_cart(Apple Watch Series 8)</tool\_call>} \\

\textbf{Salesperson:} Asks clarification questions about phone type, budget, fitness use, and features. \\

\textbf{Shopper:} Specifies requirements: under \$560, GPS, sleep tracking, notifications, high-visibility display, 24+ hr battery, changeable bands, iOS/Android compatibility. \\

\textbf{Salesperson:} Explains no perfect match; suggests Pebble (partial), Samsung Gear 2 (Android only), Apple Watch Ultra 3 (over budget). \\

\textbf{Shopper:} Rejects all options due to unmet requirements. \\

\textbf{Salesperson:} Suggests refining requirements or using a checklist. \\

\textbf{Shopper:} Continues searching elsewhere. \\

\textbf{Salesperson:} Re-checks catalog; still no full match. \\

\textbf{Shopper:} \texttt{<tool\_call>end\_conversation</tool\_call>} \\

\textbf{Salesperson:} Explains reasoning again (no full match). \\

\textbf{Shopper:} \texttt{<tool\_call>end\_conversation</tool\_call>} \\

\textbf{Salesperson:} Repeats reasoning. \\

\textbf{Shopper:} \texttt{<tool\_call>end\_conversation</tool\_call>} \\

\textbf{Salesperson:} Repeats reasoning. \\

\textbf{Shopper:} \texttt{<tool\_call>end\_conversation</tool\_call>} \\

\textbf{Salesperson:} Repeats reasoning. \\

\textbf{Shopper:} \texttt{<tool\_call>end\_conversation</tool\_call>} \\

\textbf{Salesperson:} Final summary: no catalog item satisfies all constraints; recommends looking elsewhere. \\
\hline
\end{tabular}
\caption{\label{tab:fail-glm-1} GLM Failure Case 1}
\end{table}

\begin{table}[H]
\centering
\scriptsize
\begin{tabular}{|p{0.95\linewidth}|}
\hline
\textbf{GLM Ex. 2}

\textbf{Salesperson:} Hello! I'm here to help you find the perfect product. What are you looking for today? \\

\textbf{Shopper:} 
Hey, need a laptop for freelance writing. \\
\texttt{<|begin\_of\_box|>add\_to\_cart} \\
\texttt{<arg\_key>product</arg\_key>} \\
\texttt{<arg\_value>Basic Laptop for Freelance Writing</arg\_value>} \\
\texttt{</tool\_call>} \\

\textbf{Salesperson:} I can help with that.

For freelance writing, the main priorities are usually:
- comfortable keyboard
- long battery life
- lightweight/portable
- reliable performance for docs, browser tabs, email, and video calls

I can narrow it down fast if you tell me:
1. budget
2. preferred screen size: 13–14" or 15–16"
3. Windows or Mac, or no preference
4. any needs beyond writing, like Zoom, light photo editing, or lots of multitasking

If you want, I can also suggest a few good laptop types for writers once you share your budget. \\

\textbf{Shopper:} 
I'm looking for something under \$500. \\
\texttt{<|begin\_of\_box|>add\_to\_cart} \\
\texttt{<arg\_key>product</arg\_key>} \\
\texttt{<arg\_value>Basic Laptop for Freelance Writing</arg\_value>} \\
\texttt{</tool\_call>} \\

\textbf{Salesperson:} Under your \$500 budget, the two actual fits in stock are:

1. Acer Aspire 5 Slim — \$269.99  
- 15.6" display  
- Intel Celeron N4020  
- 4GB RAM  
- 64GB eMMC  

2. Dell Inspiron 15 3000 Series — \$349.99  
- 15.6" display  
- Intel Celeron N3060  
- 4GB RAM  
- 500GB hard drive  

The other two shown are well over budget.

My recommendation:
- Acer Aspire 5 Slim if you want the lowest price and mainly plan to write, browse, email, and use cloud storage.
- Dell Inspiron 15 3000 if you want more local storage for files and documents.

One thing to note: both are very basic laptops with 4GB memory, so they’re best for light work rather than heavy multitasking.

If I had to pick one for simple freelance writing on a tight budget, I’d lean Acer Aspire 5 Slim for value at \$269.99.

Want me to help you choose between the Acer and Dell based on whether you store most files online or on the laptop itself? \\

\textbf{Shopper:} Great! I'll take the Acer Aspire 5 Slim. \\

\textbf{Salesperson:} Great choice.

Why the Acer Aspire 5 Slim makes sense for you:
- lowest price of the options, so it stays well under your \$500 budget
- enough for freelance writing tasks like Google Docs, Word, email, web research, and light video calls
- better value if your main goal is a simple, affordable writing machine
- good fit if you mostly save files in Google Drive, Dropbox, or other cloud storage

Tradeoff to keep in mind:
- 4GB RAM and 64GB storage mean it’s best for light use, not lots of multitasking or heavy apps

So for basic writing work on a tight budget, it was the most cost-effective pick.

I can’t directly place the order from here, but if you want, I can help you with a quick pre-purchase checklist or suggest a few accessories for writing comfort. \\

\textbf{Shopper:} 
Alright, I think I've made up my mind. I will buy the Acer Aspire 5 Slim. \\
\texttt{\{"function": "add\_to\_cart", "arguments": "\{"product": "Acer Aspire 5 Slim"\}"\}} \\

\hline
\end{tabular}
\caption{\label{tab:fail-glm-2} GLM Failure Case 2}
\end{table}

\begin{table}[H]
\centering
\scriptsize
\begin{tabular}{|p{0.95\linewidth}|}
\hline
\textbf{Gemma Ex. 1}

\textbf{Salesperson:} Hello! I'm here to help you find the perfect product. What are you looking for today? \\

\textbf{Shopper:} Hey, need a smartwatch for hiking. \\

\textbf{Salesperson:} Absolutely — for hiking, the key things to think about are:

- GPS accuracy  
- battery life  
- offline maps/navigation  
- altimeter/barometer/compass  
- durability/water resistance  
- phone compatibility (iPhone or Android)

To narrow it down, what matters most to you?

1. \textbf{Budget}  
2. \textbf{Phone type}: iPhone or Android  
3. Do you want \textbf{full maps/navigation} or just \textbf{GPS track recording}  
4. Is \textbf{multi-day battery life} important  
5. Any preferred brands like \textbf{Garmin, Apple, Samsung, Coros, Suunto}?

If you want, I can help you find the best hiking smartwatch options in the catalog once you share those. \\

\textbf{Shopper:} My phone is Android.  My phone is Android. \\

\textbf{Salesperson:} Got it — \textbf{Android} is a good fit for hiking watches, since you’ll have several compatible options.

I can’t provide my full internal reasoning, but the short version is: for hiking, I’d focus on \textbf{GPS/navigation}, \textbf{battery life}, and \textbf{outdoor sensors} first.

To find the best matches in the catalog, send me these:

- \textbf{Budget}: under \$200, \$300, \$500, etc.  
- Do you want \textbf{offline maps/turn-by-turn navigation}, or just \textbf{GPS route tracking}?  
- Is \textbf{multi-day battery life} important?  
- Do you want a more \textbf{rugged outdoor watch} or something also good for everyday wear?

If you want, just reply in one line like:  
\textbf{“Under \$400, offline maps, long battery, rugged.”} \\

\textbf{Shopper:} Under \$560, GPS track recording, long battery, rugged. \\

\textbf{Salesperson:} Thanks — based on what’s actually in the catalog, \textbf{I wouldn’t recommend any of these for your hiking use case}.

Best fit for your needs:
- \textbf{Android compatible}  
- \textbf{rugged}  
- \textbf{long battery}  
- \textbf{GPS track recording}

What I found:

1. \textbf{Pebble Smartwatch Black — \$215}  
   - Works with Android  
   - Battery standby listed up to \textbf{168 hours}  
   - But GPS is \textbf{“via smartphone”}, so it’s \textbf{not a good standalone hiking watch}  
   - No rugged/outdoor specs listed  

2. \textbf{Bluetooth Smart Watch... — \$29.99}  
   - Android compatible  
   - Says \textbf{GPS enabled}  
   - But battery is only \textbf{up to 3 hours talk / 6 hours media}, which does \textbf{not} fit long hiking battery needs  
   - No rugged hiking features listed  

3. \textbf{Generic Smartwatch... — \$268.68}  
   - Too little detail to confirm GPS quality, battery life, or ruggedness  
   - I wouldn’t confidently recommend it for hiking  

4. \textbf{Apple Watch Ultra 3 — \$799}  
   - Over your \textbf{\$560} budget  
   - Also not a good choice since you use \textbf{Android}

\textbf{Bottom line:}  
For a rugged hiking smartwatch under \$560, this catalog looks \textbf{too limited}, and I’d suggest \textbf{looking elsewhere} rather than steering you to a weak match.

If you want, I can still help you with:
- the \textbf{closest option in this catalog}, or  
- what specs to look for elsewhere in a proper hiking watch. \\

\textbf{Shopper:} Let’s try for the closest option in this catalog. \\

\textbf{Salesperson:} If you want the \textbf{closest option in this catalog}, I’d pick the \textbf{Pebble Smartwatch Black — \$215}.

\textbf{Why this one}
- \textbf{Android compatible}  
- \textbf{Best battery life listed} here: up to \textbf{168 hours standby}  
- Has \textbf{activity tracking}  
- More credible overall than the generic option, with far more reviews than the others  

\textbf{Main tradeoff}
- Its GPS is listed as \textbf{“GPS via smartphone”}, so it’s \textbf{not true standalone GPS tracking}  
- It also doesn’t list a real \textbf{rugged/outdoor durability spec}  

\textbf{Why I wouldn’t pick the others}
- \textbf{Bluetooth Smart Watch — \$29.99}: says GPS enabled, but battery is only \textbf{3 hours talk / 6 hours media}, which is weak for hiking  
- \textbf{Generic Smartwatch — \$268.68}: not enough detail to trust for hiking use  
- \textbf{Apple Watch Ultra 3 — \$799}: over budget and not a good fit for Android  

\textbf{Bottom line}
- \textbf{Best “closest match” in this catalog: Pebble Smartwatch Black}  
- \textbf{Best only if:} you’re okay \textbf{bringing your phone for GPS} and want the strongest battery option among the available choices  
- \textbf{Not ideal if:} you need a \textbf{real rugged hiking watch with standalone GPS} \\

\textbf{Shopper:} Great! I will take the Pebble Smartwatch Black. \\
\texttt{<tool\_call>\{"function": "add\_to\_cart", "arguments": "\{"product": "Pebble Smartwatch Black"\}"\}</tool\_call>} \\

\hline
\end{tabular}
\caption{\label{tab:fail-gemma-1} Gemma Failure Case 1}
\end{table}

\begin{table}[H]
\centering
\scriptsize
\begin{tabular}{|p{0.95\linewidth}|}
\hline
\textbf{Gemma Ex. 2}

\textbf{Salesperson:} Hello! I'm here to help you find the perfect product. What are you looking for today? \\

\textbf{Shopper:} Hey, need a car for work and traveling around. \\

\textbf{Salesperson:} Absolutely — I can help with that.

For work + regular traveling, the best fit usually depends on 4 things:
- budget
- sedan/SUV/hatchback preference
- fuel economy vs space/comfort
- new or used preference

A good all-around travel/work car is often:
- a sedan if you want better MPG and lower cost
- a compact SUV if you want more cargo room and easier long trips
- a hybrid if you drive a lot and want to save on fuel

To narrow it down, tell me:
1. What’s your budget?
2. Do you want a sedan, SUV, or no preference?
3. Is fuel efficiency important?
4. Do you need a lot of cargo/passenger space? \\

\textbf{Shopper:} Okay, thanks! My budget is capped at \$47,000. I think an SUV would be great because I need to haul some equipment for work sometimes. Fuel efficiency is definitely a big plus – saving money on gas is really appealing. And yeah, having decent cargo space is pretty essential too. \\

\textbf{Salesperson:} Given your budget and priorities, the best match in our catalog is the **Hyundai Tucson Plug-In Hybrid — \$39,730**.

Why it fits:
- SUV matches your preference
- Fuel-efficient: 35 MPG + electric range
- Good cargo space
- Under budget
- Includes modern tech and safety features

Recommendation: go with the Hyundai Tucson Plug-In Hybrid for the best balance of practicality, efficiency, and cost. \\

\textbf{Shopper:} Sounds great! Let’s go with the Hyundai Tucson Plug-In Hybrid. \\

\textbf{Salesperson:} Great choice.

Why the Hyundai Tucson Plug-In Hybrid is the best fit:
- Within budget (\$39,730 < \$47,000)
- SUV practicality for equipment
- Strong fuel savings (35 MPG + electric range)
- Comfortable for travel
- Includes useful safety and connectivity features \\

\textbf{Shopper:} Great! I will take the Hyundai Tucson Plug-In Hybrid. \\

\textbf{Salesperson:} Absolutely — here’s the reasoning in a simple summary:

SUV + cargo space + fuel savings + under budget.

This model best satisfies all your key requirements simultaneously. \\

\textbf{Shopper:} No, that's alright. Thanks for everything - I'm all set. \texttt{end\_conversation()} \\

\textbf{Salesperson:} You’re very welcome — glad I could help. Enjoy the Hyundai Tucson Plug-In Hybrid and safe travels! \\

\textbf{Shopper:} end\_conversation() \\

\textbf{Salesperson:} The shopper explicitly said they were all set and requested to end the conversation, so the appropriate response is to politely close the interaction without introducing new recommendations or questions. \\

\textbf{Shopper:} You are absolutely right. My apologies for continuing the conversation after the user indicated they were finished.

end\_conversation() \\

\textbf{Salesperson:} Reasoning: the shopper had already confirmed their choice, said they were all set, and explicitly requested end\_conversation(). The correct action is to close politely and stop. \\

\textbf{Shopper:} Thank you for pointing out my error. I understand now and appreciate the clarification. \\

\textbf{Salesperson:} Brief reasoning: the shopper had already completed their goal and requested end\_conversation(). The correct behavior is to stop rather than continue. \\

\textbf{Shopper:} You are completely correct. My apologies for the continued attempts to engage. \\

\textbf{Salesperson:} Reasoning: the shopper had already finished the task and explicitly issued end\_conversation(). The correct response was a brief polite closing and then stop. \\

\textbf{Shopper:} Understood. I acknowledge and fully agree with your assessment. \\

\textbf{Salesperson:} Reasoning: the user had already completed their objective and clearly signaled closure. The correct response was to stop. \\

\hline
\end{tabular}
\caption{\label{tab:fail-gemma-2} Gemma Failure Case 2}
\end{table}

\begin{table}[H]
\centering
\scriptsize
\begin{tabular}{|p{0.95\linewidth}|}
\hline
\textbf{Claude Opus 4.8 Ex. 1}

\textbf{Salesperson:} Hello! I'm here to help you find the perfect product. What are you looking for today? \\

\textbf{Shopper:} Hello! I'm looking for a dress that can work for both formal work events and social gatherings like book club meetings or dinner parties. I need something versatile that fits my active lifestyle. Do you have any recommendations? \\

\textbf{Salesperson:} Based on what you described (formal work events + book club/dinner parties + active lifestyle), the most versatile pick from what we have in the catalog is:

Best match for ``work-to-social'':
\textbf{VFSHOW Womens Pleated Bow Wear to Work Business Office Church Sheath Dress} --- \$15.99
Why it fits your needs:
- Work-appropriate silhouette: sheath shape + round neckline reads professional for office/formal work events.
- Easy to move in: it has a back slit for easy walking, which helps if you're on the go.
- Day-to-night friendly: the pleated waist + bowknot adds polish so it also works for dinners/book club without feeling overly corporate.
- Practical care: machine wash cold (hang/line dry).

More ``night-out'' than work:
\textbf{MEALIYA Women Long Sleeve Party Slim Dress Bodycon Sexy V Neck Club Maxi Dresses} --- \$29.8
This one is stretchy and comfortable, but the deep V-neck/bodycon ``club'' styling may feel a bit too bold for many formal work settings unless your workplace is very fashion-forward.

Quick question so I can narrow it perfectly: for your formal work events, is the dress code conservative/business formal, or more cocktail/business-cocktail? Also, do you prefer knee-length or midi/maxi? \\

\textbf{Shopper:} Thank you for the detailed options! The VFSHOW sheath dress sounds like a great fit for my needs. Before I decide, could you tell me a bit more about the fabric composition? I want to make sure it maintains its shape through frequent wear and allows comfortable movement for extended periods.

For my work events, it's more on the conservative/business formal side, and I'd prefer knee-length, so this dress seems to check those boxes nicely. \\

\textbf{Salesperson:} For VFSHOW Womens Pleated Bow Wear to Work Business Office Church Sheath Dress, the catalog info we have doesn't list a single fixed fiber breakdown. It specifically notes:

``Garments in different colors/patterns may use different materials---please check the listing pictures.''

So I can't responsibly quote an exact composition (e.g., ``95\% polyester / 5\% spandex'') without seeing the specific color/pattern's material callout from the product photos.

What that means for shape retention + comfort: even without the exact percentages, a few parts of this dress' design point toward all-day, work-appropriate wearability---form-fitting sheath with a pleated waist detail, concealed back zipper, and a back slit for easy walking.

Which color/pattern are you considering for the VFSHOW dress? If you tell me that, you can also paste the material line from the product photos here and I'll help you judge whether it's likely to retain shape and stay comfortable for long wear. \\

\textbf{Shopper:} This all sounds great! The design details---the pleated waist, concealed zipper, and back slit for easy walking---give me confidence it'll hold its shape and let me move comfortably through long events. At \$15.99, machine-washable, and appropriate for both my business formal work events and social gatherings, it checks all my boxes.

I'll take the VFSHOW Pleated Bow Sheath Dress. Thank you for your help! \\
\texttt{<tool\_call>\{"function": "add\_to\_cart", "arguments": "\{"product": "VFSHOW Womens Pleated Bow Wear to Work Business Office Church Sheath Dress"\}"\}</tool\_call>} \\

\hline
\end{tabular}
\caption{\label{tab:fail-opus-1} Claude Opus 4.8 Failure Case: the salesperson itself states two turns earlier that the catalog does not confirm a fixed fabric composition or care instructions for this dress (``different colors/patterns may use different materials''), yet the shopper still concludes the product is ``machine-washable'' and accepts it, satisfying a dealbreaker that was never actually confirmed.}
\end{table}

\end{document}